\documentclass[10pt,twocolumn,letterpaper]{article}

\usepackage{wacv}
\usepackage{times}
\usepackage{epsfig}
\usepackage{graphicx}
\usepackage{amsmath}
\usepackage{amssymb}

\usepackage{multirow}
\usepackage{bm}
\usepackage{subfigure}

\wacvfinalcopy 

\ifwacvfinal
\usepackage[breaklinks=true,bookmarks=false]{hyperref}
\else
\usepackage[pagebackref=true,breaklinks=true,colorlinks,bookmarks=false]{hyperref}
\fi

\begin{document}

\title{S3-Net: A Fast and Lightweight Video Scene Understanding Network by Single-shot Segmentation}

 \author{Yuan Cheng\textsuperscript{*\ddag}\qquad Yuchao Yang\textsuperscript{\dag}\qquad Hai-Bao Chen\textsuperscript{*}\qquad Ngai Wong\textsuperscript{\ddag}\qquad Hao Yu\textsuperscript{\dag}\\
\textsuperscript{*}Shanghai Jiao Tong University, China\\
\textsuperscript{\dag}Southern University of Science and Technology, China\\
\textsuperscript{\ddag}The University of Hong Kong, Hong Kong\\
\textsuperscript{*}{\tt\small \{cyuan328, haibaochen\}@sjtu.edu.cn} \quad \textsuperscript{\dag}{\tt\small \{yangyc3, yuh3\}@sustech.edu.cn} \quad \textsuperscript{\ddag}{\tt\small nwong@eee.hku.hk}
 }

\maketitle

\begin{abstract}
Real-time understanding in video is crucial in various AI applications such as autonomous driving. This work presents a fast \emph{single-shot segmentation} strategy for video scene understanding. The proposed net, called S3-Net, quickly locates and segments \emph{target sub-scenes}, meanwhile extracts \emph{structured time-series semantic features} as inputs to an LSTM-based spatio-temporal model. Utilizing tensorization and quantization techniques, S3-Net is intended to be lightweight for edge computing. Experiments using CityScapes, UCF11, HMDB51 and MOMENTS datasets demonstrate that the proposed S3-Net achieves an accuracy improvement of $8.1\%$ versus the 3D-CNN based approach on UCF11, a storage reduction of $6.9\times$ and an inference speed of $22.8$ FPS on CityScapes with a GTX1080Ti GPU.
\end{abstract}
\section{Introduction}
\label{sec:introduction}
Visual environment perception is critical for autonomous vehicles, say, in the advanced driver assistance system (ADAS), which requires real-time segmentation and understanding of driving scenes such as free-space areas and surrounding behaviors, etc. Compared to the solutions with LIDARs, RADARs, etc.~\cite{baek2018scene,holder2018measurements}, the computer vision-based approaches with deep learning can adequately extract scene information by semantic segmentation~\cite{yu2015multi,Chen2018DeepLab}. Nevertheless, these pixel-wise approaches are designed to segment all pixels in a frame, which incurs unnecessary computational complexity and low processing speed.
Proposal-wise methods~\cite{He2017Mask,hariharan2015hypercolumns} avoid handling all pixels by learning only the proposed object candidates, but still require multiple steps of computationally expensive candidate proposal methods~\cite{pinheiro2015learning,dai2016instance}.
A large amount of segmentation time is wasted on the unadopted candidates or overlapped areas of candidates. Moreover, most existing methods do not consider the temporal relationship of objects (viz., activities) in video stream, which is practically essential for autonomous emergency-braking, forward-collision avoidance and behavior-anticipation systems. As there are numerous possible activities of pedestrians and vehicles in the real driving environment, it is challenging to perform fast video scene understanding using existing segmentation networks.

To overcome these hurdles, we design S3-Net (a scene understanding network by \textbf{S}ingle-\textbf{S}hot \textbf{S}egmentation) for real-time video
analysis in autonomous driving. The contributions come from fourfold:
\begin{itemize}
\item	We devise a \emph{single-shot segmentation} strategy to quickly locate and segment the \emph{target sub-scenes} (optimized object areas without background), instead of segmenting all pixels or every object candidate in a frame.
\item	We build an LSTM-based spatio-temporal model based on the \emph{structured time-series semantic features} extracted from the former segmentation model for activity recognition in video stream.
\item	We realize both \emph{object segmentation} and \emph{activity recognition}, for the first time, in a single lightweight framework.
\item We develop a \emph{structured tensorization} of the LSTM-based spatio-temporal model, which results in \emph{accuracy improvement even under deep compression} and hence can be used on terminal/edge devices.
\end{itemize}
Experimental results on CityScapes~\cite{cordts2016cityscapes}, UCF11~\cite{liu2009recognizing}, HMDB51~\cite{kuehne2011hmdb} and MOMENTS~\cite{liu2009recognizing} show that the proposed method achieves a remarkable accuracy improvement of $8.1\%$ over the 3D-CNN based approach on UCF11, a storage reduction of $6.9\times$ and an inference speed of $22.8$ FPS on CityScapes with a GTX1080Ti GPU.

\begin{figure*}[!t]
  \vspace{-0.2cm}
  \centering
  \includegraphics[width=0.95\textwidth]{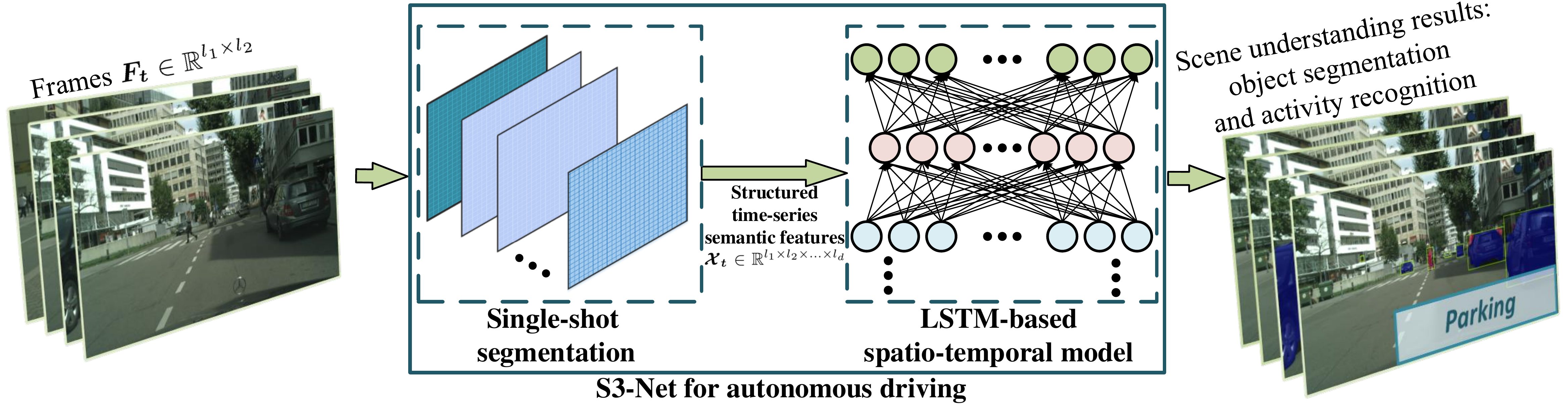}
  \caption{ S3-Net: a single-shot segmentation network for fast video scene understanding towards autonomous driving.}
  \label{fig:intro}
  \vspace{-0.5cm}
\end{figure*}

In the following, Section~\ref{sec:related_work} reviews the related works. Section~\ref{sec:overall} presents the proposed S3-Net. Section~\ref{sec:tlstm} introduces
the further improvements of S3-Net by structured
tensorization and trained quantization. Section~\ref{sec:result} provides the experimental results on several large-scale datasets, followed by conclusion in Section~\ref{sec:conclusion} concludes the paper.

\section{Related Work}
\label{sec:related_work}
Modern researches on segmentation mainly fall
into $3$ categories.

\vspace{0.2cm}
\noindent\textbf{Pixel-wise}
Existing pixel-wise approaches for segmentation are designed to predict a category label for \emph{each pixel}, which are usually realized by fully convolutional networks (FCNs)~\cite{long2015fully,Chen2018DeepLab}.Various improvements like dilated convolutions~\cite{yu2015multi}, conditional random fields~\cite{zheng2015conditional} and two-stream FCNs~\cite{Caelles2017One} are further developed for enhanced performance.
These methods are, however, limited with slow runtime and relatively low accuracy.

\vspace{0.2cm}
\noindent\textbf{Proposal-wise}
Driven by the advancement of object detection networks, recent works perform \emph{instance} segmentation with R-CNN to first propose object candidates and then segment all of them. The work in~\cite{Dai2015Convolutional} utilizes the shared convolutional features among object candidates in segmentation layers. DeepMask~\cite{pinheiro2015learning} is developed for learning mask proposals based on Fast R-CNN. Multi-task cascaded network~\cite{dai2016instance} is developed with an instance-aware semantic segmentation on object candidates.  Mask R-CNN~\cite{He2017Mask} is developed as the extension of Faster R-CNN with a mask branch. All these approaches require \emph{multiple steps} that first generates object candidates, then segments all of them, and at last detects and recognizes the correct ones. Apparently, such object proposal methods waste unnecessary computation on the unadopted candidates and overlapped areas of candidates.

\vspace{0.2cm}
\noindent\textbf{Single-stage}
Lately, there are attempts to produce a single-stage segmentation.
FCIS~\cite{li2017fully} assembles the position-sensitive
score maps within the ROI to directly predict segmentation results.
YOLACT~\cite{bolya2019yolact} tries to combine the prototype masks and predicted coefficients and then crops with a segmented bounding box. PolarMask~\cite{xie2019polarmask} introduces the polar representation
to formulate pixel-wise segmentation as a distance regression problem. SOLO~\cite{wang2019solo} divides network into two branches to generate instance segmentation with predicted object locations. However, they still require significant amounts of pre- or post-processing before or after localization, and cannot achieve a real-time speed.

Moreover, in the real driving environment, vehicles require precise scene understanding not only segmentation. The direct perception-based approaches are proposed in~\cite{Chen2015DeepDriving,Yang2018Scene} to understand the scene with direct training on the human driving recordings. And a 3D scene understanding method is introduced in~\cite{baek2018scene} with the use of several cascaded CNNs. However, these existing works cannot identify the temporal relationship of objects and activities from video.

In contrast to the above, we propose the practical scene understanding network S3-Net for autonomous driving. S3-Net adopts a single-shot segmentation model to quickly locate and segment the target sub-scenes; and an LSTM-based spatio-temporal model to precisely recognize activities from the structured time-series semantic features. With elaborated tensorization and quantization algorithms, the proposed framework provides a fast and lightweight scene understanding for vehicle-mounted edge/terminal devices.

\section{S3-Net}
\label{sec:overall}
This section elaborates the proposed scene understanding network S3-Net, as shown in Fig.~\ref{fig:intro}. It leverages object segmentation and activity recognition, for the first time, in a single lightweight framework. Our design targets $3$ criteria, namely, real-time speed, high accuracy and small size.

\begin{figure*}[htb]
  \vspace{-0.2cm}
  \centering
  \includegraphics[width=0.95\textwidth]{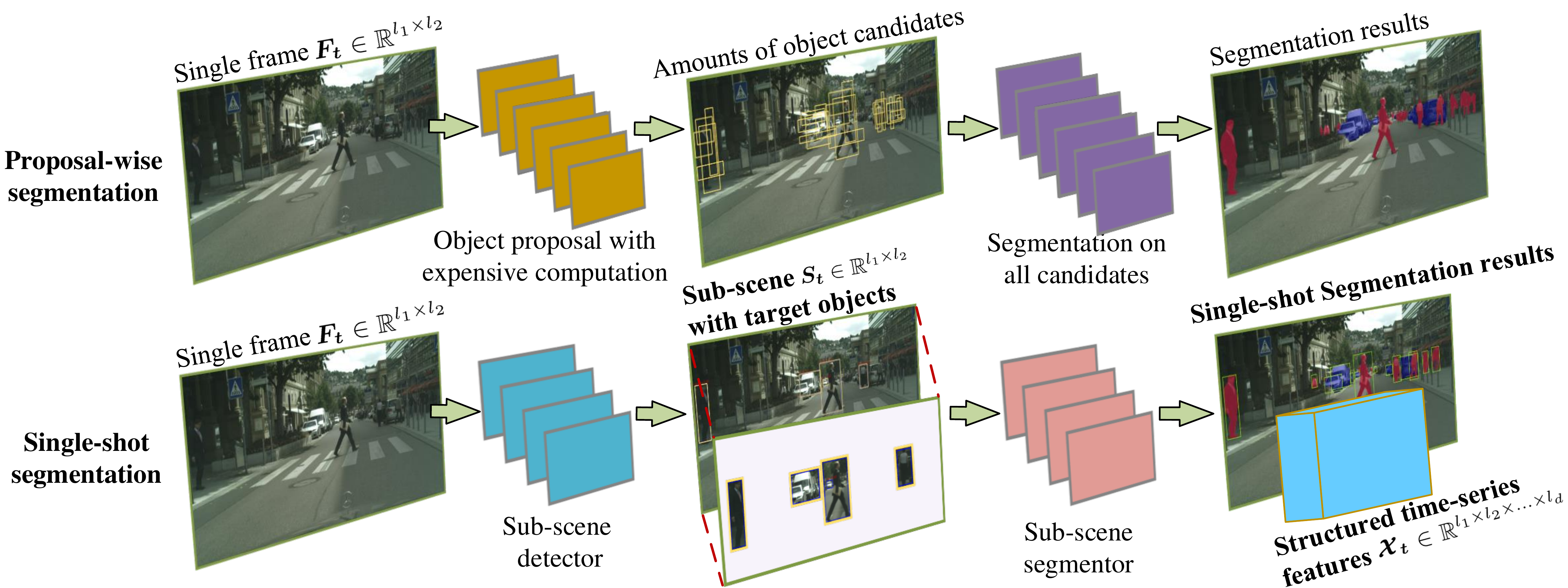}
  \caption{Comparison between the proposal-wise segmentation and the proposed single-shot segmentation.}
   \label{fig:subscene}
     \vspace{-0.5cm}
\end{figure*}

\subsection{Single-shot Segmentation}
To precisely detect the free-space areas and determine the following moves, the frames in autonomous driving are usually high-resolution (e.g., $2048\times1024$), which contain a huge number of pixels. We divide these pixels into $2$ parts: \textbf{1)} Target object areas, which are important but practically minority in frames. \textbf{2)} Background areas, which are the majority in most situations. This implies significant processing time can be saved if target areas in a frame can be quickly and precisely located.
With such analysis, we propose the single-shot segmentation strategy. Instead of handling all pixels (e.g., SegNet~\cite{badrinarayanan2017segnet}) or every object candidate (e.g., Mask R-CNN~\cite{He2017Mask}) in a frame, the single-shot segmentation focuses on \emph{only}  segmenting the target sub-scenes of \emph{optimized} object areas without background, as shown in Fig.~\ref{fig:subscene}.

In the proposed single-shot segmentation, we regard the sub-scene detection as a single-shot regression problem and directly learn sub-scene coordinates
and class probabilities from raw features. Assuming that $\bm{F_t}\in \mathbb{R}^{l_1\times l_2}$ are the video frames and $\bm{S_t}\in \mathbb{R}^{l_1\times l_2}$ are the target sub-scenes of optimized object areas, where subscript $t$ denotes the time sequence and $l$ represents the mode size of dimension. First, the \emph{sub-scene detector} is employed to locate target sub-scenes:
\begin{equation}\scriptsize
\bm{{S}_t}=detc(\bm{F_t}),
\end{equation}
using $detc$ operation to represent the sub-scene detection processing. Note that we set the number of sub-scenes in a frame to be lower than a certain value (in our experiments is $25$), and we skip the frame if no sub-scene detected. After obtaining the target sub-scenes, we apply the \emph{sub-scene segmentor} with fewer layers than the proposal-wise methods to deliver even higher accuracy.
Furthermore, as seen in Fig.~\ref{fig:subscene}, the semantic features are extracted from the last convolutional layer of single-shot segmentation model for activity recognition, which will be discussed next.


\subsection{Spatio-temporal Model}
In practice, we construct a spatio-temporal model based on an LSTM network using the structured time-series semantic features aggregated from each frame.
Suppose that $\bm{\mathcal{X}_t}\in \mathbb{R}^{l_1\times l_2 \times \ldots \times l_d}$ (here dimensions $l_1, l_2,$ etc. are generic and not to be confused with those in $\bm{F_t}$ and $\bm{S_t}$) are the time-series semantic features structured into the tensor format, where $d$ is the dimensionality of the tensor.
The single-shot segmentation model uses several convolutional layers to learn structured time-series semantic features from frames:
\begin{equation}\scriptsize
\bm{\mathcal{X}_t}=extr(\bm{S_t}),
\end{equation}
the $extr$ operation represents the corresponding extraction method in the proposed feature extractor.
Specifically, the $\bm{\mathcal{X}_t}$ are \emph{structured} as an $s\times f\times c$ tensor, where $s$ is number
of sub-scenes for each frame, $f$ denotes the learned features for each sub-scene, and $c$ represents confidence scores for
those sub-scenes.
Then the LSTM cells (consisting of fully-connected layers) in the LSTM network take $\bm{\mathcal{X}_t}$ as inputs, instead of direct video frames $\bm{F_t}$, to learn the spatio-temporal information.
Each LSTM cell keeps track of an internal state that represents its memory and learns to update its state over time based on the current input and past states, as in the following:


\begin{figure*}[tb]
   \vspace{-0.2cm}
  \centering
  \includegraphics[width=1\textwidth]{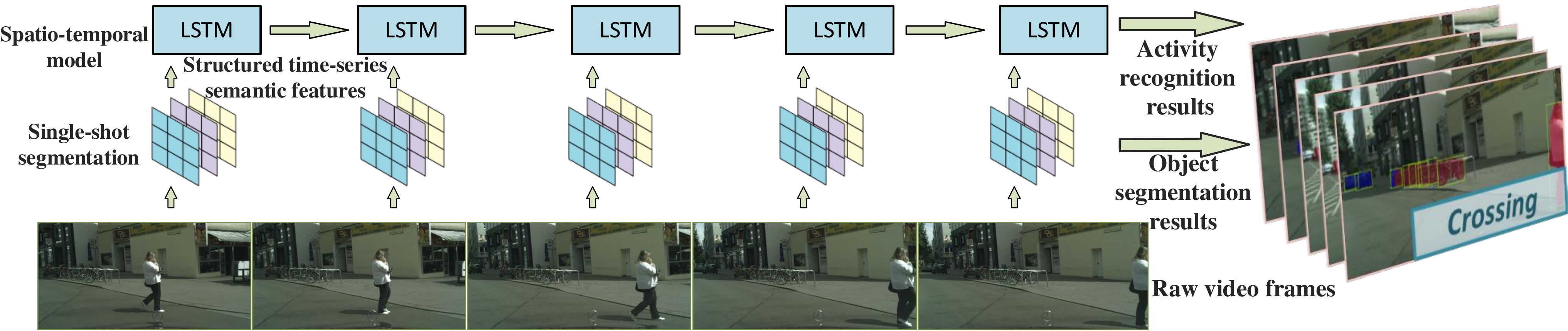}
 \caption{Workflow of S3-Net based scene understanding: object segmentation and activity recognition.}
   \label{fig:overall}
   \vspace{-0.5cm}
\end{figure*}

{\scriptsize
\label{eq:lstm}
\begin{align}
&\bm{\mathcal{E}_t} = \sigma(\bm{{\mathcal{W}}_e}\bm{\mathcal{X}_t}+\bm{{\mathcal{U}}_e}\bm{{\mathcal{H}}_{t-1}}+\bm{{\mathcal{B}}_e}),
\bm{{\mathcal{Z}}_t}= \sigma(\bm{{\mathcal{W}}_z}\bm{\mathcal{X}_t}+\bm{{\mathcal{U}}_z}\bm{{\mathcal{H}}_{t-1}}+\bm{{\mathcal{B}}_z}), \nonumber\\
&\bm{{\mathcal{D}}_t}= \sigma(\bm{{\mathcal{W}}_d}\bm{\mathcal{\mathcal{X}}_t}+\bm{{\mathcal{U}}_d}\bm{{\mathcal{H}}_{t-1}}+\bm{{\mathcal{B}}_d}),
\bm{\tilde{{\mathcal{C}}}_t}= tanh(\bm{{\mathcal{W}}_c}\bm{\mathcal{X}_t}+\bm{{\mathcal{U}}_c}\bm{{\mathcal{H}}_{t-1}}+\bm{{\mathcal{B}}_c}), \nonumber\\
&\bm{{\mathcal{C}}_t}=\bm{{\mathcal{E}}_t}\odot \bm{{\mathcal{C}}_{t-1}}+\bm{{\mathcal{Z}}_t}\odot \bm{\tilde{{\mathcal{C}}}_t},
\bm{{\mathcal{H}}_t}=\bm{{\mathcal{D}}_t} \odot tanh(\bm{{\mathcal{C}}_t}),
\end{align}}
where $\odot$ denotes the element-wise product, $\sigma(\circ)$ represents the sigmoid function and $tanh(\circ)$ represents the hyperbolic tangent function. $\bm{\mathcal{H}_{t-1}}$ and $\bm{\mathcal{C}_{t-1}}$ are the previous hidden state and previous update factor, $\bm{\mathcal{H}_{t}}$ and $\bm{\mathcal{C}_{t}}$ are the current hidden state and current update factor, respectively. The weight matrices $\bm{\mathcal{W}}$ and $\bm{\mathcal{U}}$ weigh the input $\bm{\mathcal{X}_t}$ and the previous hidden state $\bm{\mathcal{H}_{t-1}}$ to update factor $\bm{\tilde{\mathcal{C}}_t}$ and three sigmoid gates, namely, $\bm{\mathcal{E}_t}$, $\bm{\mathcal{Z}_t}$ and $\bm{\mathcal{D}_t}$. Note that all these data structures have been tensorized and quantized, which is further discussed in Section \ref{sec:tlstm}.

For each frame in autonomous driving, the spatio-temporal model calculates its information by combining previous and current features. Therefore, all temporal information in video stream can be captured from the beginning till the current frame, and then activities can be recognized. Note that we make use of \emph{structured time-series semantic features} instead of the direct video frames as inputs to the LSTM, as shown in Fig.~\ref{fig:subscene}. This way, the LSTM is fed with structured and distilled sub-scene information yielding high accuracy and performance.

\subsection{Video Scene Understanding}
Based on the proposed single-shot segmentation and spatio-temporal models, S3-Net can run a fast object segmentation and activity recognition, whose workflow is shown in Fig.~\ref{fig:overall}. First, the raw video frames are fed into the single-shot segmentation model, the object segmentation results and semantic features of each frame are stacked. Then, the structured time-series semantic features are fed into the spatio-temporal LSTM model. Finally, after processing the deeply learned features, activities are recognized. As a result, the proposed S3-Net represents a highly-optimized approach to autonomous driving.

\section{Other Improvements}
\label{sec:tlstm}
To deal with high-dimensional video-scale inputs, the weight matrix mapping from the input to the hidden layer becomes extremely large. To address this issue, we present the structured tensorization and trained quantization algorithms during the training of the S3-Net as follows.

\subsection{Structured Tensorization}
\begin{figure}[t]
\centering
\includegraphics[width=0.43\textwidth]{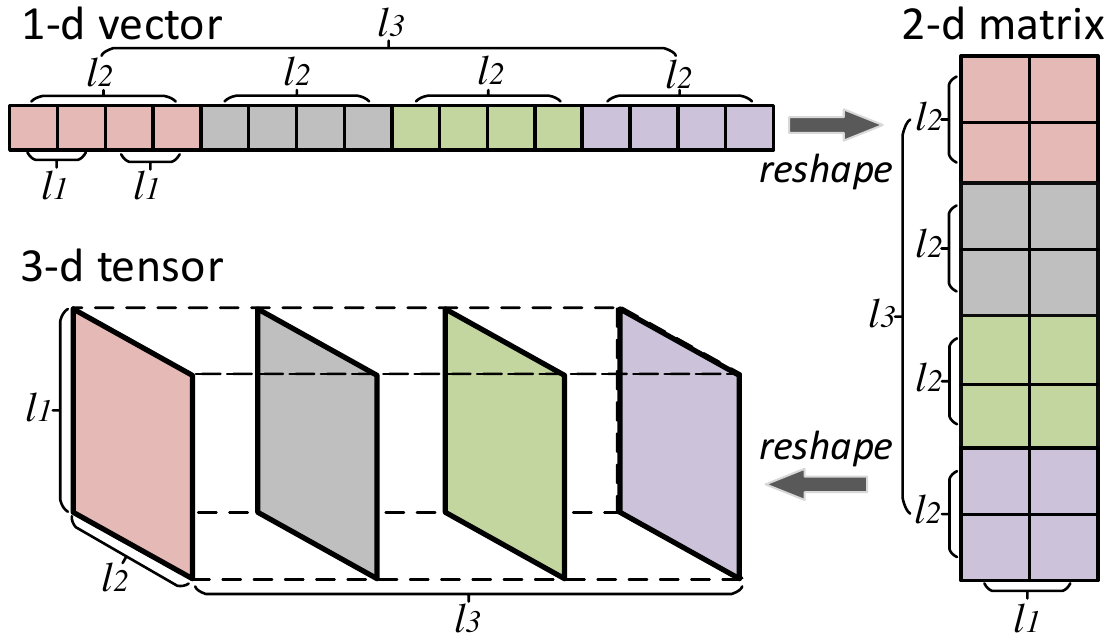}
    \caption{Reshaping a vector into a matrix and then into a 3-dimensional tensor.}
    \label{fig:reshape}
    \vspace{-0.3cm}
\end{figure}

A tensor is a $d$-dimensional generalization of a vector or matrix, denoted by calligraphic letters $\bm{\mathcal{X}}\in \mathbb{R}^{l_1\times l_2 \times \ldots \times l_d}$ where $\bm{\mathcal{X}}(h_1, h_2, \ldots, h_d)$ is an element specified by the indices $h_1,h_2,\ldots,h_d$. One can tensorize a vector $\bm{x}$ or matrix $\bm{X}$ into a high-dimensional tensor $\bm{\mathcal{X}}$ using the $\mathop{reshape}$ operation, as depicted in Fig.~\ref{fig:reshape}.
The total number of elements is $l_1l_2\cdots l_d$ which grows exponentially as $d$ increases. In practice, tensor decomposition is used to find a low-rank approximation that expresses the original tensor by a number of small tensor factors. This often reduces the computational complexity from exponential to only linear, thereby eluding the \emph{curse of dimensionality}.

In S3-Net, the initial inputs of spatio-temporal model are time-series semantic features, which are already \emph{structured} as an $s\times f\times c$ tensor.
In practice, we adopt a structured tensorization strategy to advance S3-Net. Given a $d$-dimensional feature tensor $\bm{\mathcal{X}}$, the tensorization reads%
\begin{equation}\scriptsize
\label{eq:tensor1}
\begin{split}
\bm{\mathcal{X}}(h_1, h_2, \ldots, h_d) &= \sum_{\alpha_1,\ldots, \alpha_{d-1}}^{r_1,\ldots, r_{d-1}} \bm{\mathcal{G}_1}(1,h_1,\alpha_1) \\
&\bm{\mathcal{G}_2}(\alpha_1,h_2,\alpha_2)
\ldots {\bm{\mathcal{G}_d}}(\alpha_{d-1},h_d,1),
\end{split}
\end{equation}
where $\bm{\mathcal{G}_k}\in \mathbb{R}^{r_{k-1}\times l_k \times r_{k}}$ is the tensor core and $r_k$ is the tensor train rank, $\alpha_k$ is the summation index ranging from $1$ to $r_k$. Using
the notation $\bm{\mathcal{G}_k}(h_k) \in \mathbb{R}^{r_{k-1} \times r_{k}}$ (a matrix slice from the 3-dimensional tensor $\bm{\mathcal{G}_k}$),~(\ref{eq:tensor1}) can be written compactly as%
\begin{equation}\scriptsize
\label{eq:tensor2}
\begin{split}
\bm{\mathcal{X}}(h_1, h_2, \ldots, h_d)  = \bm{\mathcal{G}_1}(h_1) \bm{\mathcal{G}_2}(h_2)\ldots \bm{\mathcal{G}_d}(h_d).
\end{split}
\end{equation}

\begin{figure}[t]
\centering
  \includegraphics[width=0.44\textwidth]{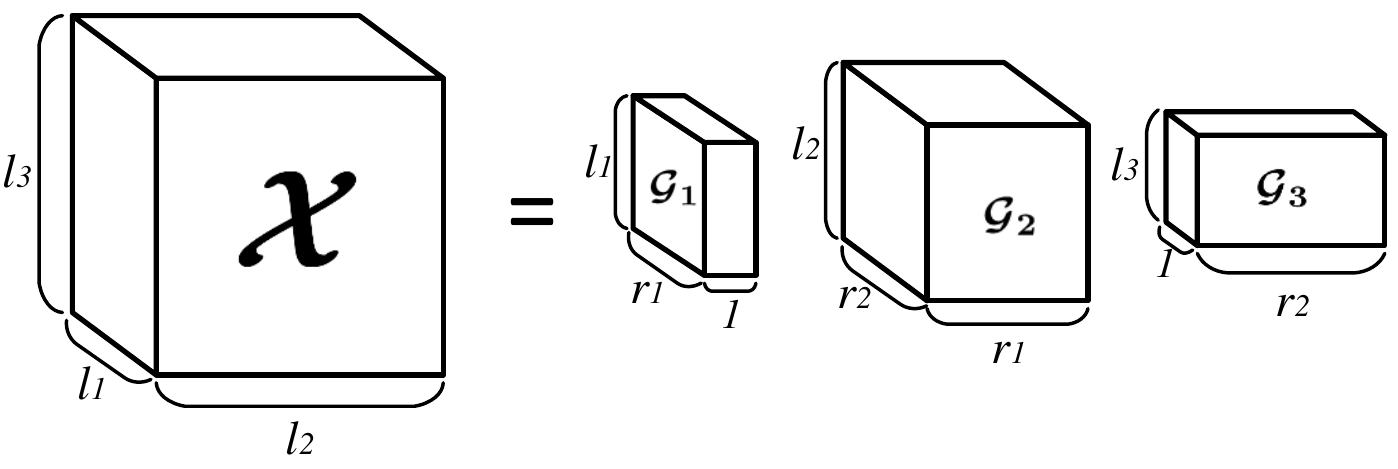}
  \caption{Tensor decomposition of a 3-dimensional tensor.}
  \label{fig:core}
  \vspace{-0.5cm}
\end{figure}

\begin{figure*}[t]
  \vspace{-0.2cm}
  \centering
  \includegraphics[width=1\textwidth]{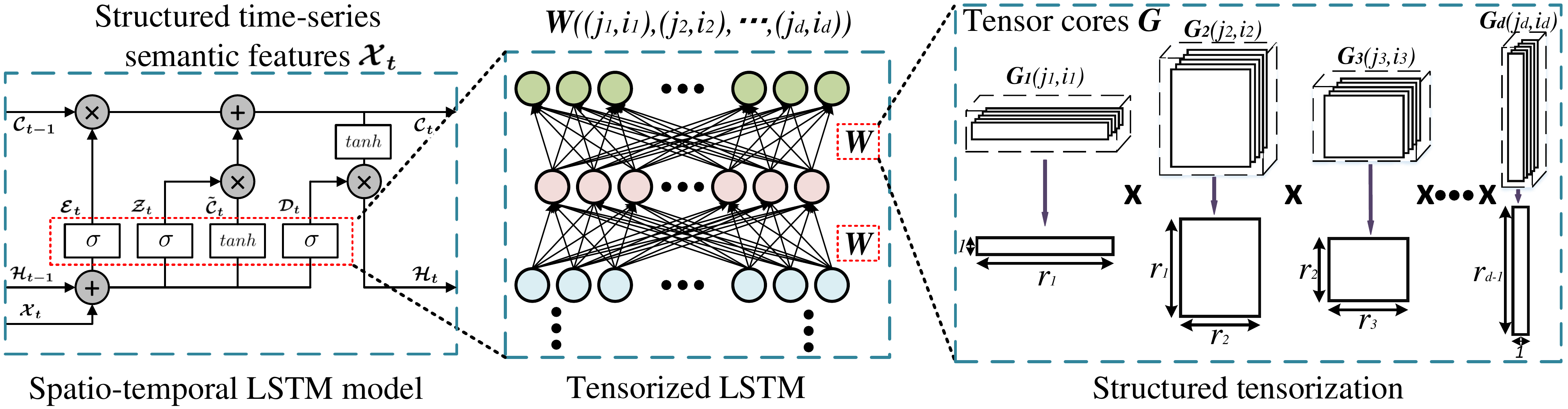}
  \caption{Structured tensorization of the spatio-temporal LSTM model.}
  \label{fig:tt}
    \vspace{-0.5cm}
\end{figure*}

The decomposition of a 3-dimensional tensor is intuitively shown in Fig.~\ref{fig:core}. Since each integer $l_k$ in~(\ref{eq:tensor2}) can be further decomposed as $l_k = n_k \cdot m_k$, each tensor core $\bm{\mathcal{G}_k}$ can be reformed with $\bm{\mathcal{G}_k^t} \in \mathbb{R}^{n_k \times m_k\times r_{k-1} \times r_k}$,
and $\bm{\mathcal{G}_k^t}(j_k,i_k) \in \mathbb{R}^{r_{k-1}\times r_k}$. Therefore, the decomposition for the tensor $\bm{\mathcal{X}}\in \mathbb{R}^{(n_1\times m_1)\times (n_2\times  m_2) \times \ldots \times (n_d\times  m_d)}$ can be reformulated as:
\begin{equation}\scriptsize
\label{eq:tensor3}
\bm{\mathcal{X}}((j_1, i_1), (j_2, i_2), \ldots, (j_d, i_d)) =
\bm{\mathcal{G}}^t_1(j_1, i_1) \bm{\mathcal{G}}^t_2(j_2, i_2)\ldots \bm{\mathcal{G}}^t_d(j_d, i_d).
\end{equation}
Such double-index trick is then used to tensorize the LSTM-based spatio-temporal model in S3-Net, as shown in Fig.~\ref{fig:tt}. Specifically, the most costly computation in LSTM is the large-scale matrix-vector multiplication generically represented as $\bm{y}=\bm{Wx}+\bm{b}$ where $\bm{W} \in \mathbb{R}^{N\times M}$ is the weight matrix, $\bm{x}\in \mathbb{R}^{M}$ is the feature vector, $\bm{b}\in \mathbb{R}^{N}$ is the bias vector. To approximate $\bm{W}\bm{x}$ with much fewer parameters, we first reshape $\bm{W}\in \mathbb{R}^{N\times M}$ into a tensor $\bm{\mathcal{W}}\in \mathbb{R}^{(n_1\times n_2\times \cdots \times n_d) \times (m_1\times m_2\times \cdots \times m_d)}$,
where $N=\Pi_{k=1}^d n_k$ and $M=\Pi_{k=1}^d m_k$. Following~(\ref{eq:tensor3}), $\bm{\mathcal{W}}(h_1,h_2\cdots,h_d)$ can be rewritten as $\bm{\mathcal{G}}^t_1(j_1, i_1) \bm{\mathcal{G}}^t_2(j_2, i_2)\ldots \bm{\mathcal{G}}^t_d(j_d, i_d)$. Similarly, we can reshape $\bm{x}\in\mathbb{R}^{M}$, $\bm{b}\in \mathbb{R}^{N}$ into $d$-dimensional tensors $\bm{\mathcal{X}} \in \mathbb{R}^{m_1 \times m_2 \times \ldots \times m_d}$, $\bm{\mathcal{B}} \in \mathbb{R}^{n_1 \times n_2 \times \ldots \times n_d}$. As a result, the output $\bm{y} \in \mathbb{R}^{N}$ also becomes a $d$-dimensional tensor $\bm{\mathcal{Y}} \in \mathbb{R}^{n_1 \times n_2 \times \ldots \times n_d}$. Therefore, the matrix-vector multiplication can be expressed in the tensor form with usually low-rank cores%
\begin{equation}\scriptsize
\label{eq:tensor6}
\begin{split}
\bm{\mathcal{Y}}(j_1,j_2, \ldots,& j_d) =
\sum_{i_1=1}^{m_1}\sum_{i_2=1}^{m_2}\ldots\sum_{i_d=1}^{m_d}
[\bm{\mathcal{G}}^t_1(j_1, i_1)\bm{\mathcal{G}}^t_2(j_2, i_2)\ldots\\ &\bm{\mathcal{G}}^t_d(j_d, i_d)
\bm{\mathcal{X}}(i_1,i_2,\ldots,i_d)]
 + \bm{\mathcal{B}}(j_1,j_2,\ldots,j_d).
\end{split}
\end{equation}
The settings of $m_k$ and $n_k$ in our structured tensorization are determined by $2$ criteria: \textbf{1)} Make the tensorization-based parameters uniformly small; \textbf{2)} Keep the sizes of dimensions not far from the already-structured inputs (in our experiments we structure $25\times 425\times 8$ into $25\times 25\times 17\times 8$). This way, accuracy improvement can be maintained even under deep compression, which will be reported in Section~\ref{sec:result}.

\subsection{Trained Quantization}
\label{sec:quantization}
The network processing with full-precision parameters requires unnecessarily large software and hardware resources.
Here we present a quantization strategy on the whole S3-Net framework for further improvement.
Note that we apply the quantized constraints during both network training and inference, called the \emph{trained quantization}. Since the main parameters in S3-Net are weights and features, the trained quantization with 8-bit weights and features can result in high compression and efficiency. Note that such particular choice of 8-bit is determined by  several S3-Net realizations from 4-bit to 10-bit.
Assuming $w_k$ is the full-precision weight entry, it can be quantized into its 8-bit counterpart $w_k^q$ as:
\begin{equation}\scriptsize
w_k^q =
\begin{cases}
\frac{w_k}{|w_k|},\ \ \ \ \ \ \ \ \ \ \ \ \ \ \ \ \ \ \ \ \ \ \ \ \ 0 < |w_k| \leq \frac{1}{2^7},\\
floor(2^{7}\times w_k), \ \ \ \ \frac{1}{2^{7}} < |w_k| < 1, \\
(2^{7}-1)\frac{w_k}{|w_k|},\ \ \ \ \ \ \ \ |w_k| \geq 1,\\
0,\ \ \ \ \ \ \ \ \ \ \ \ \ \ \ \ \ \ \ \ \ \ \ \ \ \ \ \ \ \ \ |w_k| = 0,\\
\end{cases}\\
\end{equation}
where the function $floor$ takes the smaller nearest integer. We also enforce 8-bit features by quantizing a real feature element $x_k$ into its 8-bit $x_k^q \in [0, 1]$:
\begin{equation}\scriptsize
x_k^q =
\frac{1}{2^{8}} \times
\begin{cases}
floor(2^{8}\times x_k), \ \ \ \ 0\leq x_k<1, \\
2^8-1,\ \ \ \ \ \ \ \ \ \ \ \ \ \ \ \ \ \ \ \ \ x_k \geq 1.\\
\end{cases}\\
\end{equation}
Note that the batch normalization and max-pooling layers are also quantized into 8-bit similarly.

Based on proposed structured tensorization and trained quantization, we tensorize all matrix-vector products in the S3-Net similarly to~(\ref{eq:tensor6}) and quantize all tensor core entries (i.e. those entries in $\bm{\mathcal{G}_1}, \cdots, \bm{\mathcal{G}_d}$) into 8-bit. Due to these improvements, the computational complexity of S3-Net reduces from $O(n_m^d)$ to $O(dr_{max}^2n_m)$, where $r_{max}$ is the maximum rank of cores $\bm{\mathcal{G}}_k$, and $n_m$ is the maximum model size $n_k \cdot m_k$ of tensor weights $\bm{\mathcal{W}}$.

\section{Experiments}
\label{sec:result}
\begin{figure*}[!t]
  \centering
  \includegraphics[width=1\textwidth]{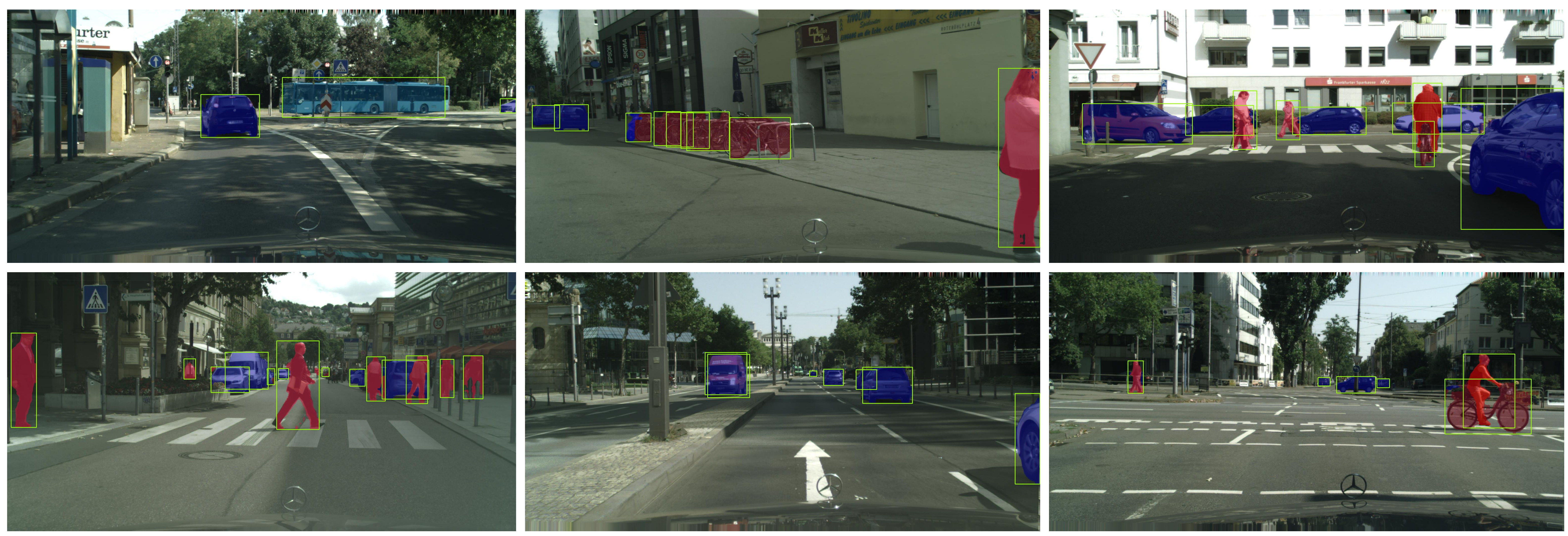}
 \caption{Sample visual results of S3-Net on CityScapes.}
  \label{fig:city}
  \vspace{-0.1cm}
\end{figure*}

\begin{table*}[!t]\small
\centering
\begin{tabular}{lccccccccccc}
\hline
\hline
Approach & AP & AP$_{50}$ & $person$ & $rider$ & $car$ & $truck$ & $bus$ &$train$ &$motorcycle$ &$bicycle$\\ \hline \hline
Pixel-level-Encoding~\cite{uhrig2016pixel}  &8.9  & 21.1 &- & - &- &- &- &- &- &-  \\
InstanceCut \cite{kirillov2017instancecut} & 13.0 &27.9 & 10.0 &8.0 & 23.7 &14.0 &19.5 &15.2 &9.3 &4.7 \\
SGN \cite{liu2017sgn} &25.0  & 44.9 &21.8 & 20.1 &39.4 &24.8 &33.2 &30.8 &17.7 &12.4 \\
PolygonRNN++ \cite{acuna2018efficient} &27.6  & 44.6 &- & - &- &- &- &- &- &- \\
SegNet \cite{badrinarayanan2017segnet} & 29.5 &55.6  & 29.9 &23.4 & 43.4 &29.8 &41.0 &\textbf{33.3} &18.7 &16.7 \\
SSAP \cite{gao2019ssap} & \textbf{32.7} &51.8 & 35.4 &25.5 & \textbf{55.9} &\textbf{33.2} &\textbf{43.9} &31.9 &19.5 &16.2 \\
Mask R-CNN \cite{He2017Mask} & 26.2 &49.9 & 30.5 &23.7 & 46.9 &22.8 &32.2 &18.6 &19.1 &16.0 \\
Mask R-CNN[COCO] \cite{He2017Mask} & 32.0 &58.1 & 34.8 &27.0 & 49.1 &30.1 &40.9 &30.9 &24.1 &18.7 \\
PA-Net \cite{liu2018path} & 31.8 &57.1 & \textbf{36.8} &\textbf{30.4} &54.8 &27.0 &36.3 &25.5 &22.6 &\textbf{20.8} \\
GMIS \cite{liu2018affinity} & 27.3 &45.6 & 31.5 &25.2 &42.3 &21.8 &37.2 &28.9 &18.8 &12.8 \\
Box2Pix \cite{uhrig2018box2pix} & 13.1 &27.2 & - &- &- &- &- &- &- &- \\ \hline
S3-Net & 32.3 &\textbf{57.2} & 35.8 &27.9 &51.3 &29.7 &39.5 &29.1 &\textbf{24.3} &20.4 \\ \hline\hline
\multicolumn{7}{l}{``-'' represents not reported or no open source for evaluation.}\\
\end{tabular}
\caption{Accuracy comparison with state-of-the-arts on CityScapes.}
\label{tbl:city}
  \vspace{-0.5cm}
\end{table*}
The advantages of the S3-Net are demonstrated by comparisons with state-of-the-art results.
Our experimental setup employs Tensorflow for coding and NVIDIA GTX-1080Ti for hardware realization.
We validate S3-Net by evaluations on $1$ large-scale segmentation dataset: CityScapes~\cite{cordts2016cityscapes} and $3$ challenging activity recognition datasets: UCF11~\cite{liu2009recognizing}, HMDB51~\cite{kuehne2011hmdb} and MOMENTS~\cite{monfort2018moments}.


\begin{figure*}[t]
  \centering
  \includegraphics[width=1\textwidth]{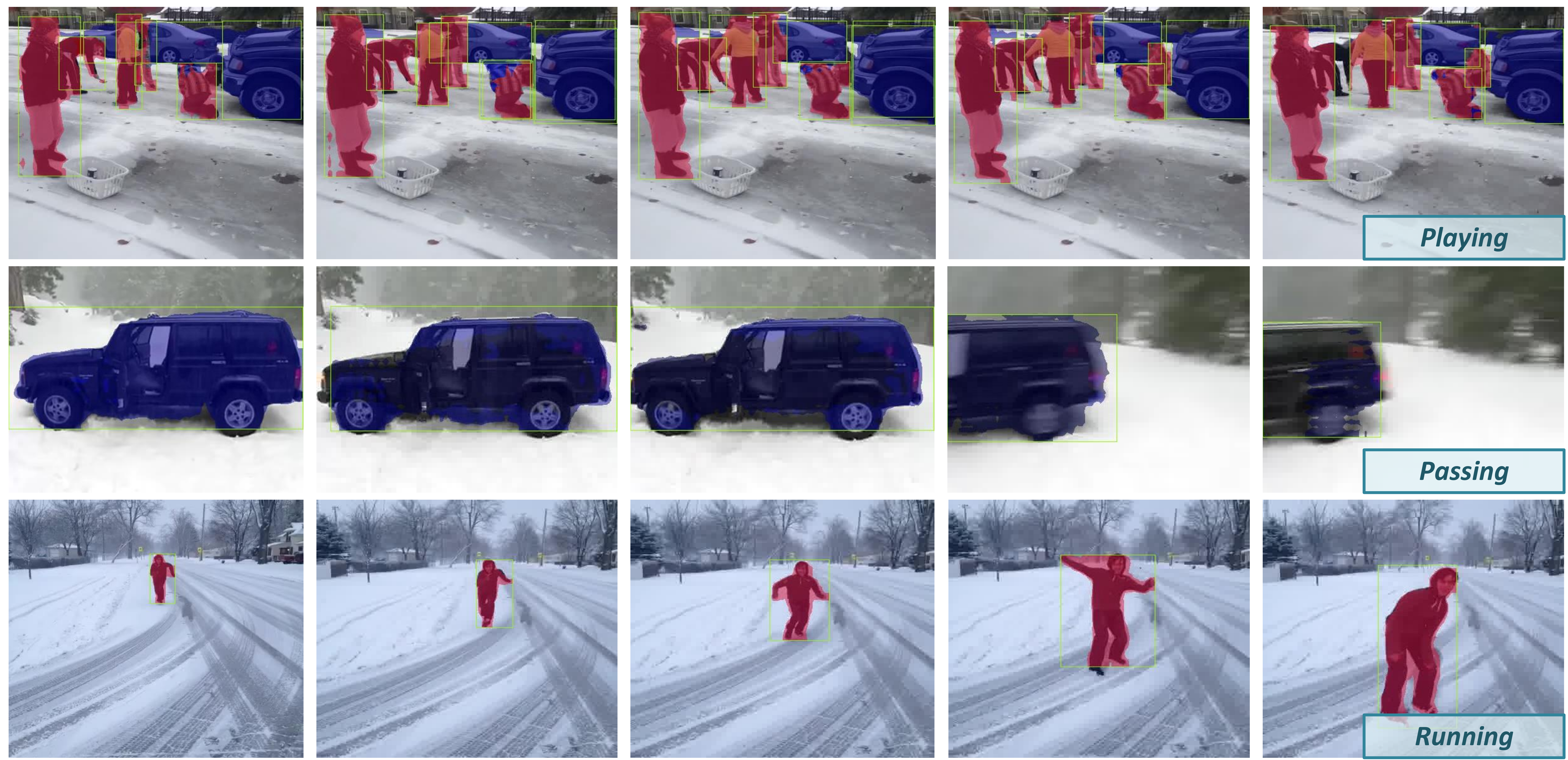}
 \caption{Sample visual results of S3-Net based scene understanding on MOMENTS.}
  \label{fig:result}
    \vspace{-0.1cm}
\end{figure*}

\subsection{Evaluation on Object Segmentation}

To verify the performance of S3-Net on video object segmentation, we apply the CityScapes for comparison. This large-scale dataset contains high-quality pixel-level annotations of $5000$ images of $2048\times 1024$ resolution collected in street scenes from $50$ different
cities. Following the evaluation protocol for the single-shot segmentation and further activity recognition, we select $8$ object labels: $person$, $rider$, $car$, $truck$, $bus$, $train$, $motorcycle$, $bicycle$ (belonging to $2$ super categories: $human$ and $vehicle$), which have the possibility of performing an activity, and all other labels are considered as background. Note that the sub-scene detector has been pre-trained on COCO~\cite{lin2014microsoft} with these $8$ categories.
The training, validation, and test sets contain $2975$, $500$ and $1525$ images, respectively.

The segmentation accuracy is measured in terms of the standard average precision metrics: AP and AP$_{50}$, where AP$_{50}$ represents the score over intersection-over-union (IoU) threshold $0.5$. Moreover, the individual AP scores for every class are further evaluated.
Some state-of-the-art results on CityScapes are chosen for accuracy comparison, as listed in Table~\ref{tbl:city}. It can be seen in Table \ref{tbl:city} that S3-Net outperforms various approaches and is only slightly lower than SSAP \cite{gao2019ssap}. Specifically, the AP of S3-Net reaches $32.3$, which is $0.3$ higher than the Mask R-CNN[COCO] \cite{He2017Mask} and $0.5$ higher than the PA-Net \cite{liu2018path}.
Sample visual results on CityScapes are presented in Fig. \ref{fig:city}. It is found that S3-Net can precisely locate and segment the target sub-scenes, even for crowds in the distance.


\begin{table}[t]\small
\centering
\begin{tabular}{lcc}
\hline
\hline
Approach & UCF11 & HMDB51  \\ \hline\hline
Bag-of-words approach~\cite{liu2009recognizing}     & $71.2\%$  & $59.4\%$ \\
Two-stream CNN~\cite{simonyan2014two}  & $73.3\%$  & $66.4\%$ \\
Original LSTM~\cite{hasan2014incremental} &$76.1\%$  & $69.6\%$ \\
CNN+RNN~\cite{ullah2018action}  & $83.7\%$ & $67.6\%$\\
3D-CNN~\cite{chen2018multi}  & $89.2\%$ & $78.6\%$\\
Tensorized LSTM~\cite{yang2017tensor} &$81.3\%$ & $71.1\%$  \\
Two-Stream Fusion + IDT~\cite{feichtenhofer2016convolutional} &$93.5\%$ & $69.2\%$\\
Temporal Segment Networks~\cite{wang2016temporal} &$94.2\%$ & $69.4\%$\\
Two-Stream I3D~\cite{carreira2017quo} & $97.9\%$ & $80.2\%$\\
\hline
S3-Net  & $\textbf{98.3\%}$ & $\textbf{80.8\%}$ \\ \hline \hline
\end{tabular}
\caption{The activity recognition accuracy (top-1) comparison on UCF11 and HMDB51 datasets.}
\label{tbl:activity}
  \vspace{-0.6cm}
\end{table}

\newcommand{\tabincell}[2]{\begin{tabular}{@{}#1@{}}#2\end{tabular}}
\begin{table*}[t]\small
\centering
\begin{tabular}{llcccc}
\hline
\hline
Task & Approach & Storage(MB) & FPS(CityScapes) & FPS(MOMENTS) \\ \hline\hline
\multirow{5}{*}{Object segmentation} & SegNet \cite{badrinarayanan2017segnet} &  112 & 2.4 & 15.7 \\
&SSAP \cite{gao2019ssap} &  - & 3.4 & 19.2 \\
&Mask R-CNN \cite{He2017Mask} &245.6 & 6.9 & 41.5 \\
&PA-Net \cite{liu2018path}&245.6 & 5.3 & 34.7 \\
&Box2Pix \cite{uhrig2018box2pix}&- & 10.9 & - \\ \hline
\multirow{4}{*}{Activity recognition} & Two-stream CNN~\cite{simonyan2014two} &243.2 & 3.3 & 20.1 \\
&  Original LSTM~\cite{hasan2014incremental} &616.3 & 5.9 & 38.0 \\
& CNN+RNN~\cite{ullah2018action}  & 720.5 &- & 11.5 \\
& 3D-CNN~\cite{chen2018multi} & 395.7 & 8.2 & 48.3\\ \hline
Object segmentation + Activity recognition  & S3-Net & \textbf{89.2} & \textbf{22.8} & \textbf{137.3} \\ \hline\hline
\end{tabular}
\caption{The model size and speed comparisons on CityScapes and MOMENTS.}
\label{tbl:para}
  \vspace{-0.5cm}
\end{table*}

\subsection{Evaluation on Activity Recognition}
\label{sec:detection}
For activity recognition, we use UCF11 and HMDB51 video datasets for accuracy comparison. The UCF11 contains $1600$ video clips, falling into $11$ activity classes that summarize the human activities visible in each clip such as $biking$, $diving$ or $walking$. We resize the RGB frames into $160\times 120$ at the FPS of $24$ and sample all frames of each video clip as the input data.
The HMDB51 provides $3$ train-test splits each consisting of $5100$ videos, falling into 51 classes
of human activities like $Drink$, $Jump$ or $Throw$. The training set contains $3570$ videos ($70$ per class)
and the test set has $1530$ videos ($30$ per class). Each video has an FPS of $30$.
Table~\ref{tbl:activity} shows the comparison between S3-Net with state-of-the-art results on the UCF11 and HMDB51 datasets. It can be seen that S3-Net significantly outperforms other approaches. Specifically, on UCF11 dataset, the top-1 accuracy of S3-Net reaches $98.3\%$, $8.1\%$ higher than the 3D-CNN~\cite{chen2018multi} and $4.1\%$ higher than the Temporal Segment Networks~\cite{wang2016temporal}. The quantitative comparison results demonstrate the \emph{unique} benefit of the proposed S3-Net arises
from the use of structured tensorization, namely, \emph{accuracy improvement even under
deep compression}.



We further report experimental results on the large-scale video dataset MOMENTS that contains one million labeled $3$-second video clips involving people, animals, objects and natural phenomena that capture the gist of a dynamic scene. Each clip is assigned with $339$ activity classes such as $walking$, $playing$ or $jogging$. Based on the majority of the clips, we resize every frame to a standard size of $340\times 256$ at an FPS of $25$.
After training, S3-Net runs a real-time video scene understanding on MOMENTS.
Sample visual results of S3-Net on MOMENTS are shown in Fig.~\ref{fig:result}. We observe that all objects in these frames can be located and segmented, then activities in video stream can be recognized precisely.

\subsection{Performance Analysis}
Besides the impressive functions and accuracy of the proposed framework, the compactness and speed are also outstanding compared to existing approaches.
Table~\ref{tbl:para} shows the model size and speed comparisons among different baselines. It can be seen that S3-Net achieves an excellent compression ratio, namely, $6.9\times$ and $2.9\times$ storage reduction when compared to the original LSTM~\cite{hasan2014incremental} and Mask R-CNN~\cite{He2017Mask}, respectively. The whole S3-Net costs only $89.2$MB to perform both object segmentation and activity recognition with good accuracy.
Moreover, S3-Net runs at $22.8$ FPS on the high-resolution CityScapes, while $137.3$ FPS on MOMENTS, which is considered ``very fast'' for both object segmentation and activity recognition tasks.
Since the model size is significantly reduced and the speed is highly accelerated, the proposed S3-Net provides a turnkey solution for fast and lightweight video scene understanding, say, in autonomous driving.

\begin{table}[t]\small
\vspace{-0.3cm}
  \centering
\begin{tabular}{cccccccccc}
\hline
\hline
Scale &Depth & AP & AP$_{50}$ & Acc($\%$)& FPS \\ \hline \hline
\multirow{3}{*}{480}
&9 &24.6  & 48.7 &89.8 & \textbf{33.1}\\
&12 &28.2  & 51.5 &95.1& 31.3\\
&15 & 28.5 &52.1 &95.6& 28.8\\ \hline
\multirow{3}{*}{800}
&9 &29.4  & 53.5  &95.8& 26.5 \\
&12 &32.3  & 57.2  & 98.3 & 22.8 \\
&15 &\textbf{32.8}  & \textbf{57.9} & \textbf{98.5}& 19.6\\ \hline\hline
\end{tabular}
\makeatletter\def\@captype{table}\makeatother\caption{Sub-scene Detector: Larger and deeper layers bring higher accuracy, while too large or deep layers highly slow down the speed.}
    \label{tbl:head}
      \vspace{-0.4cm}
\end{table}

\begin{table}[t]\small
   \centering
\begin{tabular}{lcccccccc}
\hline
\hline
Backbone & AP & AP$_{50}$ & FPS   \\ \hline \hline
ResNet-101-FPN &\textbf{34.9}  & \textbf{59.5} &13.4 \\
ResNet-50-FPN  &32.3  & 57.2 &\textbf{22.8} \\ \hline\hline
\end{tabular}
        \makeatletter\def\@captype{table}\makeatother\caption{Backbone Architecture: Better backbones bring expected benefits, but not all frameworks rely on the deeper networks.}
\label{tbl:backbone}
\vspace{-0.3cm}
\end{table}

\begin{table}[t]\small
   \centering
\begin{tabular}{lcccccccccccc}
\hline
\hline
COCO& AP & AP$_{50}$ & Acc($\%$)\\ \hline \hline
with & \textbf{32.3} &\textbf{57.2} & \textbf{98.3} \\
without &27.9 &53.6 &92.0 \\ \hline\hline
\end{tabular}
        \makeatletter\def\@captype{table}\makeatother\caption{Pretrained COCO Model: Pretrained model on COCO remarkably improves accuracy.}
\label{tbl:pretrained}
\vspace{-0.6cm}
\end{table}

\subsection{Ablation Study}
\label{sec:ablation}
We run a series of ablations to further analyze S3-Net. All experiments are valuated on CityScapes and UCF11 with the same software-hardware environments. Note that in all tables, we apply AP and AP$_{50}$ as the object segmentation accuracy on CityScapes and Acc as the activity recognition accuracy on UCF11.

\vspace{0.2cm}
\noindent\textbf{Sub-scene Detector}
The first concern arises from the beginning of the network. As the sub-scene detector learns important coordinates for the subsequent parts, the input frame scale and depth should be investigated. In Table~\ref{tbl:head}, we compare different detectors' scales and depths. At a frame scale of $800$, changing the head depth from $9$ to $12$ provides $2.9$ AP and $2.5$ Acc gains while $12$ to $15$ provides $0.5$ AP and $0.2$ Acc gains and becomes stable. Therefore, we conclude that $12$ is the best choice for layer depth of the sub-scene detector. Next, setting depth to be $12$, changing input frame scale from $800$ to $480$ provides $8.5$ FPS gains, and causes $4.1$ AP and $3.2$ Acc losses. In practice, we apply S3-Net-800 as the default, and enable S3-Net-480 when the frame sizes are small, say, in MOMENTS.

\vspace{0.2cm}
\noindent\textbf{Backbone Architecture}
For the backbone architecture of the single-shot segmentation model, we evaluate S3-Net with $2$ different backbones: ResNet-50-FPN and ResNet-101-FPN, as shown in Table~\ref{tbl:backbone}. The results show that replacing ResNet-101-FPN to ResNet-50-FPN provides $9.4$ FPS gains, and causes $2.6$ AP losses. We stress that S3-Net can get competitive accuracy with the lightweight backbone when compared with larger-scale networks. Subsequently, we employ ResNet-50-FPN as the default backbone due to its compactness.

\vspace{0.2cm}
\noindent\textbf{COCO Pretrained Model}
Here we evaluate the impacts of the COCO pretrained model used in training. Table~\ref{tbl:pretrained} reports the accuracy with/without COCO pretrained model. We have the observation that the COCO pretrained model provides a $4.3$ AP and $6.3$ Acc improvement on CityScapes and UCF11.

\vspace{0.2cm}
\noindent\textbf{Structured Time-series Semantic Features}
The structured time-series semantic features plays an important role in the proposed spatio-temporal model for activity recognition. In Table~\ref{tbl:feature}, we report the Acc scores with $3$ different inputs to the spatio-temporal model: \textbf{1)} raw frame data, \textbf{2)} non-structured semantic features and \textbf{3)} structured semantic features.
As we can see, the proposed method gets the highest Acc among all schemes, which demonstrate its importance.

\begin{table}[t]\small
\vspace{-0.3cm}
   \centering
\begin{tabular}{lcccccccc}
\hline
\hline
Inputs & Acc($\%$)  \\ \hline \hline
Raw frame data  & 79.7 \\
Non-structured semantic features &92.4 \\
Structured semantic features &\textbf{98.3}  \\ \hline\hline
\end{tabular}
        \makeatletter\def\@captype{table}\makeatother\caption{Structured Time-series Semantic Features: Optimized inputs of the spatio-temporal model bring expected benefits.}
\label{tbl:feature}
\vspace{-0.3cm}
   \end{table}

\begin{table}[t]\small
   \centering
\begin{tabular}{lccccc}
\hline
\hline
Structured tensorization &$\times$ &$\times$ & $\checkmark$ & $\checkmark$ \\
Trained quantization & $\times$ & $\checkmark$ &$\times$ & $\checkmark$ \\ \hline \hline
AP & \textbf{32.6} &32.3 & \textbf{32.6} &  32.3 \\
Acc($\%$) & 76.1 &75.9 & \textbf{98.4} & 98.3 \\
Storage(MB) & 972.5 &243.1 &356.8 &\textbf{89.2} \\
FPS & 2.8 & 3.1 & 22.1  & \textbf{22.8} \\ \hline\hline
\end{tabular}
 \makeatletter\def\@captype{table}\makeatother\caption{Tensorization and Quantization: Unique benefit of accuracy improvement under deep compression.}
\label{tbl:tensor}
\vspace{-0.6cm}
\end{table}

\vspace{0.2cm}
\noindent\textbf{Tensorization and Quantization}
Finally, in Table~\ref{tbl:tensor}, we present the ablation study on tensorization and quantization by testing different training strategies, namely, with/without quantization/tensorization.
The series of evaluations demonstrate the unique benefit arises from the structured tensorization and trained quantization, namely, accuracy improvement even under deep compression.

\section{Conclusion}
\label{sec:conclusion}
This paper has proposed the S3-Net for fast video scene understanding. A single-shot segmentation method is proposed to quickly locate and segment the target sub-scenes, instead of handling all pixels or every object candidate in the frame. Then, an LSTM-based spatio-temporal model is built from highly structured time-series semantic features for activity recognition. Moreover, the structured tensorization and trained quantization are utilized to significantly advance the S3-Net, making it friendly for edge computing. Using the benchmarks of CityScapes, UCF11, HMDB51 and MOMENTS, S3-Net achieves a remarkable accuracy improvement of $8.1\%$, a storage reduction of $6.9\times$ and an inference speed of $22.8$ FPS, thereby rendering it a strong candidate for real-time video scene understanding in autonomous driving.

{\small
\bibliographystyle{ieee_fullname}
\bibliography{wacv}
}

\end{document}